\documentclass[conference]{IEEEtran}
\IEEEoverridecommandlockouts
\usepackage{amssymb,amsfonts}
\usepackage{times}
\usepackage{enumitem}   
\usepackage{bbm}
\usepackage{lipsum}
\usepackage{comment}
\usepackage{array}
\usepackage{mathtools} 
\usepackage{algorithm}
\usepackage{amsmath,amsthm,bm}
\usepackage{soul,xcolor}
\usepackage{mdwmath}
\usepackage{mdwtab}
\usepackage{blindtext}
\usepackage{mathrsfs}
\usepackage{subcaption}
\usepackage{amsmath,amssymb,amsfonts}
\usepackage{algorithmic}
\usepackage{graphicx}
\usepackage{textcomp}
\usepackage{amsmath}

\usepackage[utf8]{inputenc}
\usepackage[english]{babel}
\newtheorem{theorem}{Theorem}
\newtheorem{corollary}{Corollary}

\newtheorem{lemma}{Lemma}
\newtheorem{definition}{Definition}

\def\BibTeX{{\rm B\kern-.05em{\sc i\kern-.025em b}\kern-.08em
    T\kern-.1667em\lower.7ex\hbox{E}\kern-.125emX}}
\begin{document}

\title{Trading Data For Learning: Incentive Mechanism For On-Device Federated Learning
}

\author{\IEEEauthorblockN{Rui Hu, Yanmin Gong}
\IEEEauthorblockA{Department of Electrical and Computer Engineering, University of Texas at San Antonio, San Antonio, TX 78249}
}
\maketitle

\begin{abstract}
Federated Learning rests on the notion of training a global model distributedly on various devices. Under this setting, users' devices perform computations on their own data and then share the results with the cloud server to update the global model. A fundamental issue in such systems is to effectively incentivize user participation. The users suffer from privacy leakage of their local data during the federated model training process. Without well-designed incentives, self-interested users will be unwilling to participate in federated learning tasks and contribute their private data. To bridge this gap, in this paper, we adopt the game theory to design an effective incentive mechanism, which selects users that are most likely to provide reliable data and compensates for their costs of privacy leakage. We formulate our problem as a two-stage Stackelberg game and solve the game's equilibrium. Effectiveness of the proposed mechanism is demonstrated by extensive simulations.
\end{abstract}

\section{Introduction}
With the growing popularity of machine learning, it is expected that the data-driven intelligent applications will soon be employed in all aspects of our daily life, including medical care, food and agriculture, transportation systems, etc.
In traditional machine learning methods, the key of training an accurate model is to collect a sufficient amount of data, which may contain private information about individuals.
If such data is disclosed or used for other purposes other than those initially intended, individual's privacy will be compromised. Indeed, data privacy is emerging as one of the most serious concerns of machine learning. Many data owners are reluctant to share their private data for the purpose of machine learning. To promote private data circulation, data brokers such as Acxiom \cite{databroker} have emerged to bridge the gap between data owners and data consumers. Basically, the data brokers offer monetary rewards to incentivize data owners to contribute private data and then charge data consumers for their queries over the collected data \cite{niu2019making}. This practice, however, has two fundamental issues: 1) data owners have no control of data privacy after transferring private data to the data broker; 2) the data broker has to take full responsibility of protecting users' data which is costly and may damage the reputation of the data broker if data breach occurs. 

Recently, federated learning has attracted increasing attentions due to its significant advantages in privacy protection. It unleashes a new collaborative ecosystem in machine learning to train a global model while keeping the training data locally on users' devices. The participating devices send the model updates computed on their raw data to a cloud server iteratively to update the global model. Comparing with the data broker systems, federated learning systems only consist of a cloud server (i.e., the data consumer) and a number of devices/users (i.e., the data owners). The cloud server has no control of users' raw data but only collects intermediate model updates from users, which contain much less sensitive information than raw data. Since data never leaves users' devices, the cloud server has no responsibility to maintain and protect the user's raw data. Therefore, federated learning is a promising tool for training machine learning models on private data.

Federated learning successfully mitigates users' concerns over privacy leakage by allowing devices to keep their data locally and only exchange ephemeral model updates.
However, it still has privacy issues \cite{kairouz2019advances}. For example, by observing the model updates from a device, attackers are able to recover the private dataset in that device using the reconstruction attack \cite{al2016reconstruction} or infer whether a sample is in the dataset of that device using the membership inference attack \cite{shokri2017membership}. Especially, if the server is not fully trusted, it can easily infer the private information of users from the received model updates during the training by employing existing attack methods. Considering such risks, self-interested devices/users will be unwilling to participate in federated learning tasks.

To motivate users with sensitive data to participate in federated learning tasks, the server should provide rigorous privacy guarantees for participants. 
Recent studies have specifically focused on solving the privacy issues in distributed learning scenarios. 
Among them, secure multi-party computation or homomorphic encryption is one of the popular methods which prevents attackers outside the system from obtaining the local computation results \cite{MoZh17}. However, these methods cannot prevent the privacy leakage from the final learned model. Besides, differential privacy \cite{dwork2014algorithmic} has become the de-facto standard for privacy notion and is being increasingly adopted in private distributed learning systems (see \cite{guo2018practical,huang2019dp,abadi2016deep} and references therein). 
However, most of these existing works made an optimistic assumption that there are enough users who are willing to participate in federated learning when invited, which is not practical due to the privacy concern of users. Without well-designed economic reward, users will be reluctant to join the learning. Therefore, it is essential for the server to design an efficient incentive mechanism to attract more user participation. There are several papers that have studied the incentive design for federated learning considering the communication and computation cost of users \cite{kang2019incentive,pandey2020crowdsourcing}, but none of them consider the privacy issue.

In this paper, we propose a game-theory based incentive mechanism to motivate users with private data to participate in federated learning tasks with rigorous privacy guarantee. In our mechanism, the server compensates users for contributing their private data, according to their privacy budgets. Users who have a larger privacy budget for the federated learning task will get higher payment from the server. Each user selects its desired privacy budget to maximize its own utility, and the server selects the reward for users such that its utility is maximized.

The main contributions of this paper are listed as follows:
\begin{itemize}
    \item To the best of our knowledge, we are the first to study the incentive mechanism that motivates users with private data to participate in federated learning tasks.
    \item Our proposed incentive mechanism features the properties of differential privacy to quantify the privacy loss and compensates participants in a satisfying manner.
    \item We perform extensive simulations to demonstrate the effectiveness of our incentive mechanism.
\end{itemize}

The remainder of this paper is organized as follows. In Section~\ref{sys-model}, we describe the data trading system for on-device federated learning tasks. Then, we formulate the utilities of the server and users in federated learning tasks in Section~\ref{pro-formulate}. In Section~\ref{solution}, we present our incentive mechanism based on the game theory. Numerical results are presented in Section~\ref{sec:eva} followed by the conclusions in Section~\ref{sec:con}.

\section{System Modeling}\label{sys-model}

In this section, we first describe the basic architecture of trading private data for federated learning tasks. Then, we present more details on conducting the on-device federated training.

\subsection{Data Trading Process}

\begin{figure}[t]
\centering
\includegraphics[width=3in]{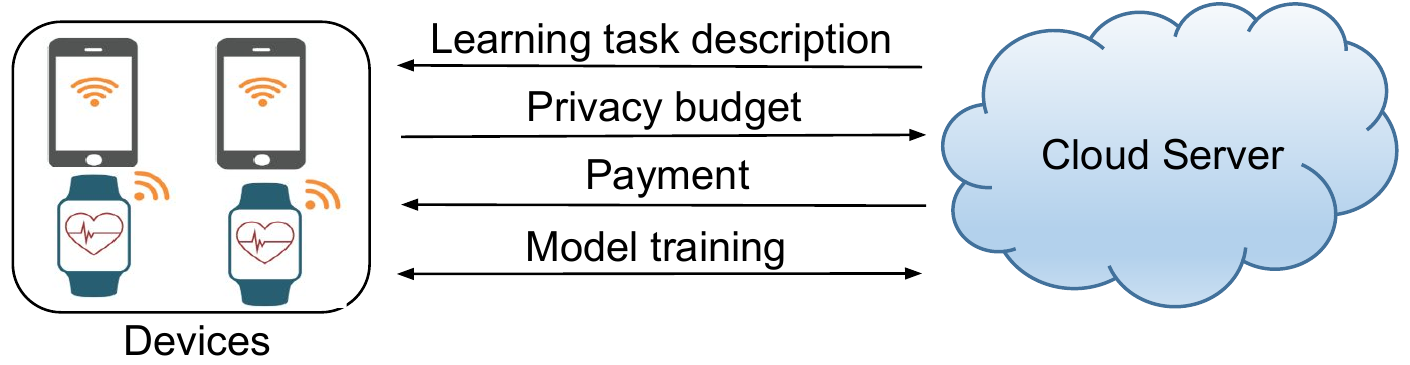}
\caption{Data trading system for federated learning tasks.}
\label{fig:data_trading}
\vspace*{-15pt}
\end{figure}

We use Figure~\ref{fig:data_trading} to aid our description of the data trading system for federated learning. The system consists of a cloud server, which aims to learn a machine learning model $\bm{\theta}\in\mathbb{R}^d$, and many devices which are able to communicate with the cloud server. We assume that these devices are owned by different users. The server first publicizes the federated learning task description. Assume that there is set of users $\mathcal{U}=\{1,2,\dots,n\}$ interested in joining the task after reading the task description, where $n\geq2$. Each user in $\mathcal{U}$ has a local dataset $D_i, i\in\mathcal{U}$. A user participating in the task will incur a privacy loss to be elaborated later. Therefore, it expects a payment in return for its service. Taking the privacy loss and payment into consideration, each user determines how much privacy budget it will give to this learning task and submits its plan to the server. After receiving all the privacy budget plans from users, the server computes the payment for each user and sends the payments to the users. The chosen users (whose payment is positive) will conduct the federated model training process. This completes the whole process of trading private data for federated learning tasks. In the following, we further specify the details of each step.

\subsubsection{Learning task description} 
At the very beginning, the cloud server broadcasts the description of a federated learning task to all users. The description includes: 1) the goal of this task, e.g., classify different pets; 2) the category of the data needed for the task, e.g., images, videos or voices of pets; 3) the loss function $f$ to be used; 4) the type of query $\mathcal{Q}$ for users, e.g., the gradient of loss computed on local data; 5) the privacy compensation function for users; 6) the total reward $R>0$ for all participants.

\subsubsection{Privacy budget}
According to the task description, each user decides its plan of privacy budget. The $(\epsilon, \delta)$-differential privacy (DP) \cite{dwork2014algorithmic} is a commonly-used concept to quantify the privacy loss in private machine learning algorithms \cite{hu2020personalized,hu2020cpfed}. Here, instead of directly using the $(\epsilon, \delta)$-DP, we utilize its relaxed version, the $\rho$-zero-concentrated differential privacy ($\rho$-zCDP), which has a tight composition bound and is more suitable to analyze the end-to-end privacy loss of iterative algorithms. Based on the concept of $\rho$-zCDP, we assume that user $i$ chooses its privacy budget $\rho_i > 0$ for this learning task. Larger $\rho_i$ implies more privacy loss. In the following, we provide several important properties of $\rho$-zCDP \cite{bun2016concentrated}:
\begin{lemma}\label{rho-zcdp}
Let $g:x\rightarrow\mathbb{R}$ be any real-valued function with sensitivity $\Delta_2(g) $, then the Gaussian mechanism, which returns $g(x) + \mathcal{N}(0,\sigma^2)$, satisfies $\Delta_2(g)^2/(2\sigma^2)$-zCDP.
\end{lemma}
\begin{lemma}\label{composition}
Suppose two mechanisms satisfy $\rho_1$-zCDP and $\rho_2$-zCDP, then their composition satisfies $\rho_1+\rho_2$-zCDP.
\end{lemma}

\subsubsection{Payment}
After collecting all the privacy budgets from users, the server computes the payment for each user based on the total reward and privacy compensation function, which is denoted by $p_i$. In the rest of this paper, we utilize the privacy compensation function which assigns payments to users proportionally to their privacy budgets, i.e., 
\begin{equation}
\label{pri_com}
    p_i = \frac{\rho_i}{\sum_{i\in\mathcal{U}} \rho_i}R.
\end{equation}
We can observe that the payment of a user depends on not only the total reward $R$ but also the privacy budgets of other users.

\subsubsection{Model training}
After the user received its payment, it will start the federated model training process under the coordination of the server. The federated training is an iterative process. At each iteration, the server sends a query to each user, and each user computes the received query on its local data. To protect their privacy, users perturb their computation results by adding random noise before sending them to the server. More precisely, at iteration $t$, user $i$ receives the query $\mathcal{Q}(\bm{\theta}^t;D_i)$ where $\bm{\theta}^t$ represents the latest model parameter maintained by the server and $D_i$ represents the dataset of user $i$.
Then, user $i$ computes the query and upload a differentially-private response to the server, i.e.,
\begin{equation}
    \tilde{\mathcal{Q}}(\bm{\theta}^t;D_i) =  \mathcal{Q}(\bm{\theta}^t;D_i) + \mathbf{b}_i^t,
\end{equation}
where $\mathbf{b}_i^t \in \mathbb{R}^d $ is a random vector drawn from the Gaussian distribution $\mathcal{N}(0, \sigma_i^2\mathbf{1}_d)$. This process repeats for $T$ iterations. The noise magnitude $\sigma_i$ depends on user $i$'s privacy budget and the number of iterations and is determined by the server. 

\subsection{Details on Private Federated Learning}
Consider a federated learning setting that consists of a cloud server and a set of users $\mathcal{U}$ which are able to communicate with the server. Each user has a local dataset $D_i=\{\xi_1^i, \dots, \xi_m^i\}$, a collection of $m$ datapoints from its device. The users want to collaboratively learn a global model $\bm{\theta}$ using their data under the orchestration of the cloud server. 
Specifically, the global model $\bm{\theta}$ is learned by minimizing the overall empirical risk of the loss on the union of all local datasets, that is,
\begin{equation}
\label{fed_obj}
    \min_{\bm{\theta}} f(\bm{\theta}) := \frac{1}{|\mathcal{U}|}\sum_{i\in\mathcal{U}}  f_{i}(\bm{\theta}) \text{ with } f_{i}(\bm{\theta}) := \frac{1}{m}\sum_{\xi\in{D}_i} l(\bm{\theta}, \xi).
\end{equation}
Here, $f_i(\cdot)$ represents the local objective function of user $i$, $l(\bm{\theta};\xi)$ is the loss of the model $\bm{\theta} $ at a datapoint $\xi$ sampled from local dataset $D_i$.

In the traditional distributed gradient descent approach that solves Problem~\eqref{fed_obj}, the server collects the gradients of local objectives from all users and updates the global model using a gradient descent iteration given by
\begin{equation}
\label{traditional_GD}
    \bm{\theta}^{t+1} = \bm{\theta}^{t} -\frac{\eta }{|\mathcal{U}|}\sum_{i\in\mathcal{U}} \nabla f_i(\bm{\theta}^{t}),
\end{equation}
where $\bm{\theta}^{t} $ represents the global model at iteration $t$, $\eta$ is the stepsize, and $\nabla f_i(\bm{\theta}^{t}) := \frac{1}{m}\sum_{\xi\in D_i}\nabla l(\bm{\theta}^t,\xi)$ represents the gradient of local objective function $f_i$ based on the local dataset $D_i$. In Algorithm~\ref{algorithm-1}, we summarize the process of training the model $\bm{\theta}$ in a private manner.
\begin{algorithm}[ht]
\caption{Private Federated Learning Algorithm}
\label{algorithm-1}
\begin{algorithmic}[1]
\REQUIRE number of iterations $T$, stepsize $\eta$, noise magnitude $\sigma_i$
\FOR{$t=0$ to $T-1$}
    \FOR{all users in $\mathcal{U}$ in parallel}
        \STATE Download the query $\mathcal{Q}(\bm{\theta}^t;D_i) = \nabla f_i(\bm{\theta}^{t}) $;
        \STATE Return the DP-response $\tilde{\mathcal{Q}}(\bm{\theta}^t;D_i)$ to the server;
    \ENDFOR
    \STATE Server updates $ \bm{\theta}^{t+1} \leftarrow \bm{\theta}^{t} - \frac{\eta }{|\mathcal{U}|}\sum_{i\in\mathcal{U}} \tilde{\mathcal{Q}}(\bm{\theta}^t;D_i)$.
    \ENDFOR
\end{algorithmic}
\end{algorithm}

\section{Utilities of The Server and Users}\label{pro-formulate}
At the beginning of the data trading process for federated learning tasks, the server needs to determine the total reward $R$ for this task. Since users are selfish but rational, they will select their privacy budgets to maximize their own utilities and will not join in the learning task unless there is sufficient incentive. The server is only interested in maximizing its own utility, hence our goal is to design a mechanism for the server to choose the best strategy considering users' decisions during the data trading process. In this section, we formulate the utilities of the server and users, respectively.

\subsection{Utility of User}
In the learning task description, the server announces a total reward $R$ to motivate users. Then, each user will decide its level of participation, i.e., its privacy budget. The goal of each user is to determine the optimal privacy budget $\rho_i$ that maximizes its utility. Denote the privacy cost of user $i$ as a function of $c(\nu_i, \rho_i)$, where $\nu_i > 0$ is the privacy value parameter. According to the privacy compensation function~\eqref{pri_com}, the utility of user $i$ can be formulated as:
\begin{equation}
\label{user_u}
    U_i = \frac{\rho_i}{\sum_{i\in\mathcal{U}} \rho_i}R - c(\nu_i, \rho_i).
\end{equation}
Note that, each user's privacy cost function belongs to the same publicly known family, but the privacy value parameter $\nu_i$ is known only to the user and must be reported to the mechanism. We require that the family of cost functions admits a total ordering independently of $\rho_i$, i.e., for any $ j\neq k$ and for any $\rho_i>0$, it should hold that $ c(\nu_i^j, \rho_i) \leq c(\nu_i^k, \rho_i)$ if and only if $\nu_i^j\leq \nu_i^k$. Natural choices of such privacy cost functions can be linear functions which take the form $c(\nu_i, \rho_i) =  \nu_i\rho_i$, exponential functions of the form $c(\nu_i, \rho_i) = \exp(\nu_i\rho_i)$, and quadratic cost functions of the form $c(\nu_i, \rho_i) =  \nu_i\rho_i^2$. In this paper, we assume all users use the linear privacy cost function.

\subsection{Utility of Server}
The goal of the server is to choose the optimal reward $R$ that maximizes its utility. The utility of the server relies on the training result, i.e., the performance of the trained model, and the reward it paid to participants. In our system, only the privacy budgets will impact the model performance via the Gaussian noise added to the model parameter at each iteration. Therefore, we are only interested in capturing the influence of users' privacy budgets on the model performance. It is impossible to obtain the exact accuracy of a trained model before conducting the training. Instead, we analyze the influence of the privacy budget on the convergence property of the federated learning task, which implies the expected performance of the trained model. 

We first observe the convergence property of Algorithm~\ref{algorithm-1}. Assume that global loss function $f$ is $L$-smooth, so that the loss gap between two iterations is
\begin{equation}
   f(\bm{\theta}^{t+1}) - f(\bm{\theta}^{t}) \leq \langle \nabla f(\bm{\theta}^{t}), \bm{\theta}^{t+1}- \bm{\theta}^{t}\rangle + \frac{L}{2}\| \bm{\theta}^{t+1}- \bm{\theta}^{t}\|^2.
\end{equation}
Taking the expectation of the loss gap over the Gaussian noise, we have
\begin{multline}
\label{variance_noise}
    \mathbb{E}\left[f(\bm{\theta}^{k+1}) - f(\bm{\theta}^{k})\right] \leq 
    -\frac{\eta}{|\mathcal{U}|}\sum_{i\in\mathcal{U}}
    \langle \nabla f(\bm{\theta}^{k}),  \mathcal{Q}(\bm{\theta}^t;D_i)\rangle \\
     +  \frac{L\eta^2}{2|\mathcal{U}|}\sum_{i\in\mathcal{U}}
    \mathbb{E}\left[\left\|\mathcal{Q}(\bm{\theta}^t;D_i) + \mathbf{b}_i^t\right\|^2\right].
\end{multline}
As we have that
\begin{align}
   \nonumber
   &\mathbb{E}\left[\left\|\mathcal{Q}(\bm{\theta}^t;D_i) + \mathbf{b}_i^t\right\|^2\right] \\
    \nonumber&= 
    \mathbb{E}\left[\left\|\mathcal{Q}(\bm{\theta}^t;D_i) + \mathbf{b}_i^t - \mathbb{E}\left[\mathcal{Q}(\bm{\theta}^t;D_i) + \mathbf{b}_i^t\right]\right\|^2\right]\\
    \nonumber&\ \ + \left\|\mathbb{E}\left[\mathcal{Q}(\bm{\theta}^t;D_i) + \mathbf{b}_i^t\right]\right\|^2 \\
    \nonumber& = \mathbb{E}\left[\left\|\mathbf{b}_i^t\right\|^2\right] + \left\|\mathcal{Q}(\bm{\theta}^t;D_i)\right\|^2,
\end{align}
the expectation of the loss gap becomes
\begin{multline}
\label{exp_gap}
    \mathbb{E}\left[f(\bm{\theta}^{k+1}) - f(\bm{\theta}^{k})\right] \leq -
    \frac{\eta}{|\mathcal{U}|}\sum_{i\in\mathcal{U} }
    \langle \nabla f(\bm{\theta}^{k}),  \mathcal{Q}(\bm{\theta}^t;D_i)\rangle \\
    + \frac{L\eta^2}{2|\mathcal{U}|}\sum_{i\in\mathcal{U} }\left\|\mathcal{Q}(\bm{\theta}^t;D_i)\right\|^2 + \frac{L\eta^2}{2|\mathcal{U}|}\sum_{i\in\mathcal{U} } \mathbb{E}\left[\left\|\mathbf{b}_i^t\right\|^2\right].
\end{multline}

From \eqref{exp_gap}, we can see that the Gaussian noise adds extra error on the loss at each iteration proportional to the size of noise. Such errors will accumulate as more and more iterations are involved. Indeed, with a fixed number of iterations, larger loss error per iteration implies lower accuracy of the trained model. If the error caused by the Gaussian noise is zero, i.e., no noise is added to the local response, Algorithm~\ref{algorithm-1} will achieve the highest accuracy approaching 1. As the magnitude of the Gaussian noise increases, the accuracy will drop to 0.5, which is the probability of random guess. Given that $\mathbb{E}\|\mathbf{b}_i^t\|^2 = d\sigma_i^2$, the utility of the server can be formulated as follows:
\begin{equation}
\label{server_u_1}
    {U}_s = \frac{\lambda}{2} \left[1+\exp{\left(-\frac{1}{|\mathcal{U}|}\sum_{i\in\mathcal{U} }\log(1+{d\eta^2\sigma_i^2})\right)}\right] - R.
\end{equation}
Here, $\lambda > 1$ is the weight parameter. The first term represents the influence of Gaussian noise on model accuracy. Specifically, its inner log term reflects the diminishing influence of the noise, and its outer term bounds the accuracy in the range of $[0.5,1]$. 

In the following, we determine the magnitude of Gaussian noise for each user. 
By Lemma~\ref{rho-zcdp}, the magnitude of noise $\sigma_i$ is proportional to the sensitivity of the query function. Therefore, we first compute the sensitivity of the query at each iteration, which is given in Corollary~\ref{sens}.
\begin{corollary}
\label{sens}
The sensitivity of the query $\mathcal{Q}(\bm{\theta}^t;D_i)$ for user $i$ at the $t$-th iteration is bounded by ${2L}/{m}$.
\end{corollary}
\begin{IEEEproof}
For user $i$, given any two neighboring datasets ${D}_i$ and ${D}_i^\prime$ of size $m$ that differ only in the $j$-th data sample, the sensitivity of the query  $\mathcal{Q}(\bm{\theta}^t;D_i)$ is
\begin{multline*}
\left\|\frac{1}{m}\sum_{\xi\in D_i}\nabla l(\bm{\theta}^t,\xi) - \frac{1}{m}\sum_{\xi\in D_i^\prime}\nabla l(\bm{\theta}^t,\xi)\right\|_2 \\
= \frac{1}{m} \left\|\nabla l(\bm{\theta}_i^{t}; \xi_j ) - \nabla l(\bm{\theta}_i^{t}; \xi_j^\prime) \right\|_2. 
\end{multline*}
Since the loss function $l(\cdot)$ is $L$-smooth, the sensitivity of $\mathcal{Q}(\bm{\theta}^t;D_i)$ is bounded by $2L/m$.
\end{IEEEproof}

Given the sensitivity of the query and the privacy budget, the server is able to calculate the magnitude of noise for each user by Theorem~\ref{thm:privacy}.
\begin{theorem}
\label{thm:privacy}
Algorithm~\ref{algorithm-1} achieves $\rho_i$-zCDP for user $i$ if the Gaussian noise $\mathbf{b}_i^{t}$ at each iteration is sampled from $ \mathcal{N}(0,\sigma_i^2\mathbf{1}_d)$, where
\begin{equation}
    \sigma_i = \frac{L}{m}\sqrt{\frac{2T}{\rho_i}}.
\end{equation}
\end{theorem}
\begin{IEEEproof}
By Corollary~\ref{sens} and Lemma~\ref{rho-zcdp}, each iteration of Algorithm~\ref{algorithm-1} achieves $ \frac{2L^2}{ m^2\sigma_i^2}$-zCDP for user $i$. By Lemma~\ref{composition}, the overall zCDP guarantee for user $i$ after $T$ iterations is $\rho_i = \frac{2TL^2}{ m^2\sigma_i^2}$. Theorem~\ref{thm:privacy} follows by simple rearrangement.
\end{IEEEproof}

Then, the server's utility function \eqref{server_u_1} can be rewritten as
\begin{equation}
\label{server_u}
    {U}_s = \frac{\lambda}{2} \left[1+\exp{\left(-\frac{1}{|\mathcal{U}|}\sum_{i\in\mathcal{U} }\log(1+\frac{2d\eta^2T}{\rho_i m^2})\right)}\right] - R,
\end{equation}
where we can see that the privacy budget $\rho_i$ is proportional to the model accuracy.

\section{Incentive Mechanism: A Two-Stage Stackelberg Game}\label{solution}
As we mentioned, the goal of the server and users is to maximize their utilities. In this section, we model the incentive mechanism as a two-stage Stackelberg game. Specifically, in the first stage, the server announces its reward $R$; in the second stage, each user strategizes its privacy budget to maximize its own utility. Let $\rho=\{\rho_1,\rho_2 ,\dots, \rho_n\}$ denote the strategy set consisting of all users' strategies. Let $\rho_{-i}$ denote the strategy set excluding $\rho_i$. 

The second stage can be considered as a non-cooperative game. Therefore, given a reward $R$, there exists a stable strategy for each user such that a user has nothing to gain by unilaterally changing its current strategy which corresponds to the concept of Nash Equilibrium in game theory.

\begin{definition}[Nash Equilibrium]
\label{nash}
A set of strategies $\rho^e=\{\rho_1^e,\rho_2^e, \dots, \rho_n^e\}$ is a Nash Equilibrium of the second stage in our Stackelberg Game if for any user $i$,
\begin{equation}
    {U}_i(\rho_i^e,\rho_{-i}^e) \geq {U}_i(\rho_i,\rho_{-i}^e) 
\end{equation}
for any $\rho_i>0$, where ${U}_i$ is defined in \eqref{user_u}.
\end{definition}

Given $ \rho_{-i}$, if a strategy maximizes $ {U}_i(\rho_i,\rho_{-i}) $ over all $ \rho_i> 0$, it is user $i$'s best strategy, denoted by $\beta_i(\rho_{-i})$. To study the best strategy, we compute the derivatives of $ {U}_i $ with respect to $ \rho_i$:
\begin{equation}
\label{first_de}
    \frac{\partial {U}_i}{\partial \rho_i} = \frac{-R \rho_i}{(\sum_{j\in\mathcal{U}}\rho_j)^2} +\frac{R}{\sum_{j\in\mathcal{U}}\rho_j} -\nu_i,
\end{equation}
and
\begin{equation}
\label{second_i}
    \frac{\partial^2 {U}_i}{\partial \rho_i^2} = \frac{-2R\sum_{j\in\mathcal{U}\backslash \{i\}}\rho_j}{(\sum_{j\in\mathcal{U}}\rho_j)^3}< 0.
\end{equation}
We can see that the utility function $U_i$ is a concave function of $\rho_i$. Therefore, given any $R>0$ and any strategy profile $\rho_{-i}$ of other users, the best strategy $\beta_i(\rho_{-i}) $ of user $i$ is unique if it exists. By setting \eqref{first_de} to zero, $\beta_i(\rho_{-i}) $ satisfies that 
\begin{equation}
\label{best_st}
    \beta_i(\rho_{-i}) =
    \begin{cases}
    -\infty, \ \text{if } R\leq \nu_i \sum_{j\in\mathcal{U}\backslash \{i\}}\rho_j;\\
    \sqrt{\frac{R \sum_{j\in \mathcal{U}\backslash \{i\}}\rho_j}{\nu_i}} - \sum_{j\in \mathcal{U}\backslash \{i\}}\rho_j, \ \text{o.w.}
    \end{cases}.
\end{equation}
Here, if the best strategy $\beta_i(\rho_{-i}) $ is non-positive, user $i$ will not participate in the training by setting $\rho_i=-\infty$ (to avoid a deficit). The conclusion in \eqref{best_st} leads to the following algorithm for computing the Nash Equilibrium of the second-stage in our game.

\begin{algorithm}[ht]
\caption{Computation of the Nash Equilibrium}
\label{algorithm-2}
\begin{algorithmic}[1]
\STATE Sort users according to their privacy value, $\nu_1\leq\nu_2\leq \dots\leq\nu_n$;
\STATE $ \mathcal{S}\leftarrow\{1,2\}, i\leftarrow3$;
\WHILE{$i\leq n$ and $\nu_i \leq \frac{\nu_i + \sum_{j\in\mathcal{S}} \nu_j}{|\mathcal{S}|}$}
    \STATE $ \mathcal{S}\leftarrow \mathcal{S} \cup \{i\}, i\leftarrow i+1$;
\ENDWHILE
\FOR{$i\in[n]$}
    \IF{$i\in\mathcal{S}$} 
        \STATE $\rho_i^e=\frac{(|\mathcal{S}|-1)R}{\sum_{j\in\mathcal{S}} \nu_j}\left(1-\frac{(|\mathcal{S}|-1)\nu_i}{\sum_{j\in\mathcal{S}} \nu_j}\right) $;
    \ELSE
        \STATE $\rho_i^e=-\infty$;
    \ENDIF
\ENDFOR
\RETURN $\rho^e=\{\rho_1^e,\rho_2^e, \dots, \rho_n^e\}$
\end{algorithmic}
\end{algorithm}

Based on Algorithm~\ref{algorithm-2}, we have the following observations: 1) $\nu_i \geq \frac{\sum_{j\in\mathcal{S}} \nu_j}{|\mathcal{S}|-1}$, for any $i\notin\mathcal{S}$; 2) $\sum_{j\in\mathcal{S}} \rho_j^e = \frac{(|\mathcal{S}|-1) R}{\sum_{j\in\mathcal{S}} \nu_j}$; and 3) $\sum_{j\in\mathcal{S}\backslash \{i\}} \rho_j^e = \frac{(|\mathcal{S}|-1)^2 R \nu_i}{(\sum_{j\in\mathcal{S}} \nu_j)^2}$ for any $i\in\mathcal{S}$. According to 1) and 2), we can obtain that for any $i\notin\mathcal{S}$, $\rho_i^e = -\infty$ is its best strategy given $\rho_{i-}^e $, which satisfies the condition in \eqref{best_st}. Then, it is provable that for any $i\in\mathcal{S}$, $\rho_i^e$ is its best strategy given $\rho_{i-}^e $. The detail of the proof is similar to the proof of Theorem~1 in \cite{yang2012crowdsourcing}. Accordingly, by Algorithm~\ref{algorithm-2}, we can obtain the best strategy profile for users, i.e.,
\begin{equation}
\label{best_i}
    \beta_i(\rho_{-i}) = \frac{(|\mathcal{S}|-1)R}{\sum_{j\in\mathcal{S}} \nu_j}\left(1-\frac{(|\mathcal{S}|-1)\nu_i}{\sum_{j\in\mathcal{S}} \nu_j}\right)
\end{equation}
if $i\in\mathcal{S}$, and $\beta_i(\rho_{-i}) =-\infty$ if $i\notin\mathcal{S} $.

On the basis of the above analysis, the server knows that there exists a unique Nash Equilibrium for users for any given value of $R$. Hence, by choosing an optimal $R$, the server is able to maximize its utility $U_s$. Substituting \eqref{best_i} into \eqref{server_u} and considering $\rho_i=-\infty$ if $ i\notin \mathcal{S}$, we have
\begin{equation}
\label{u_s_update}
    {U}_s = \frac{\lambda}{2} \left[1+\exp{\left(-\frac{1}{|\mathcal{S}|}\sum_{i\in\mathcal{S} }\log(1+\frac{1}{X_iR})\right)}\right] - R,
\end{equation}
with
\begin{equation}
    X_i = \frac{m^2(|\mathcal{S}|-1)}{2d\eta^2 T\sum_{j\in\mathcal{S}} \nu_j}\left(1-\frac{(|\mathcal{S}|-1)\nu_i}{\sum_{j\in\mathcal{S}} \nu_j}\right).
\end{equation}
Taking the second order derivative of ${U}_s $ with respect to $R$, we have
\begin{align*}
    \frac{\partial^2 {U}_s }{\partial R^2} & = \frac{\lambda g}{2R^2}\left[\left(\frac{1}{|\mathcal{S}|}\sum_{i\in\mathcal{S}} \frac{1}{X_iR + 1}\right)^2 - \frac{1}{|\mathcal{S}|}\sum_{i\in\mathcal{S}} \frac{2X_iR +1}{(X_iR + 1)^2} \right]\\
    &\leq \frac{\lambda g}{2R^2}\left(\frac{1}{|\mathcal{S}|}\sum_{i\in\mathcal{S}} \frac{- 2X_iR}{(X_iR + 1)^2}\right) < 0,
\end{align*}
where
\begin{equation*}
    g = \exp{-\left(\frac{1}{|\mathcal{S}|}\sum_{i\in\mathcal{S} }\log(1+\frac{1}{X_iR})\right)}.
\end{equation*}
Therefore, the utility function of the server in \eqref{u_s_update} is strictly concave with respect to $R$ for $R>0$. Since the value of $U_s$ in \eqref{u_s_update} approaches 0 when $R$ approaches 0 and goes to $-\infty$ when $R$ goes to $\infty$, there exists a unique maximizer $R^*$ that can be computed through either bisection or Newton's method.

\section{Numerical Evaluation}\label{sec:eva}
In this section, we evaluate the performance of our incentive mechanism in various scenarios. We assume that the privacy value parameter of each user is uniformly distributed over $[1, \nu_{max}]$, where $\nu_{max}$ is the maximum privacy value. We set system parameter $\lambda = 20$, model dimension $d=1000$, stepsize $\eta=0.1$, number of iterations $T=500$, and data size $ m=1000$. We take the number of participating users and the utilities of the server and users as our evaluation metrics. We mainly conduct simulations to study the impact of the number of users and the range of privacy value on these three evaluation metrics. We use the maximum privacy value $\nu_{max}$ to represent the range of privacy value. 
In all simulations that study the impact of the number of users $n$, we set $\nu_{max}=5$ and vary the value of $n$ from 100 to 1000. In all simulations that study the impact of the range of privacy value, we set $n=1000$ and vary the value of $\nu_{max}$ from 2 to 10.

\subsection{Number of participants}
\begin{figure}[ht]
\begin{subfigure}[t]{0.5\linewidth}
  \centering
  \includegraphics[width=\linewidth]{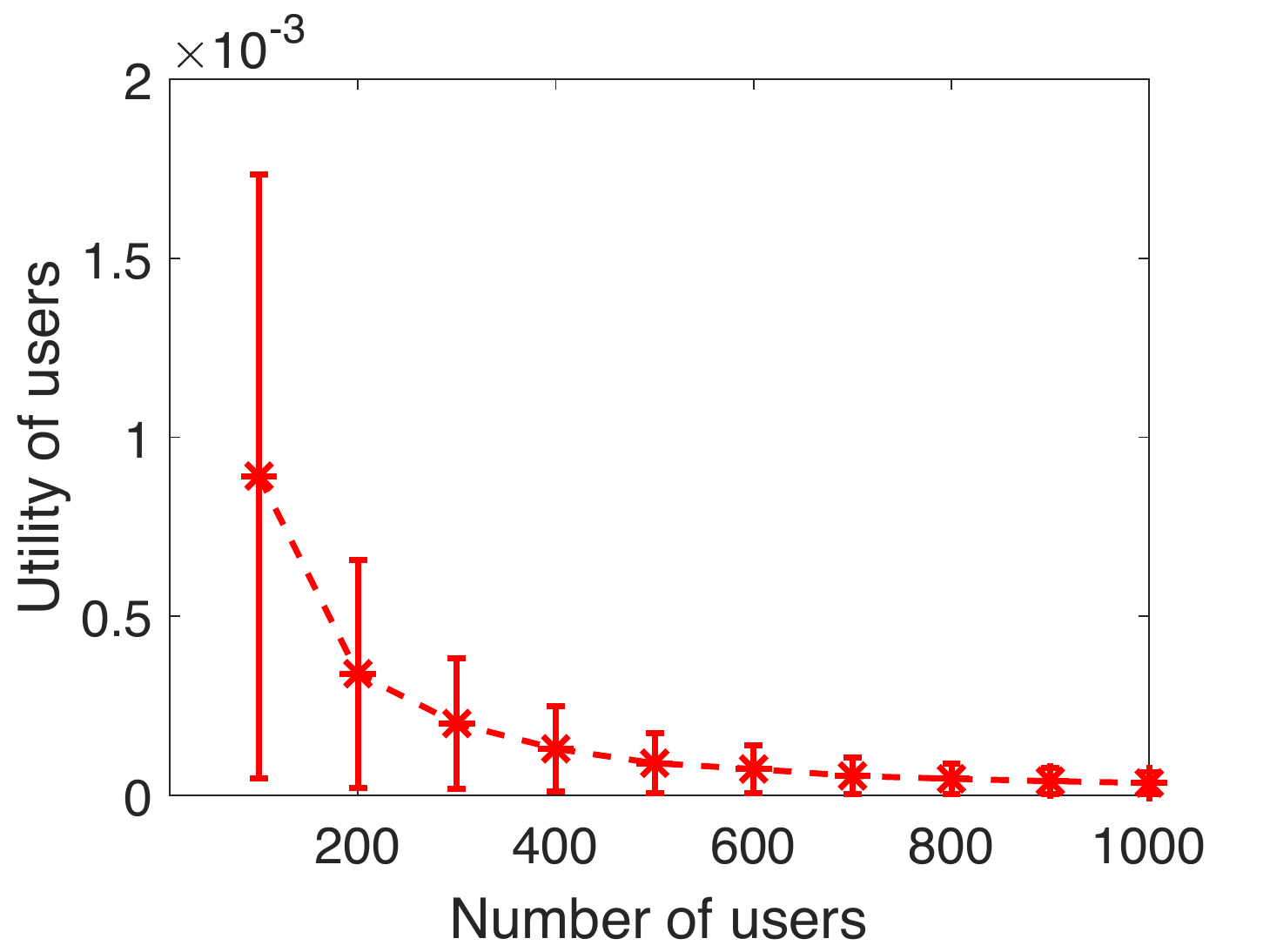}
  \caption{Impact of $n$.}
  \label{fig:num_user_num}
\end{subfigure}%
\begin{subfigure}[t]{0.5\linewidth}
  \centering
  \includegraphics[width=\linewidth]{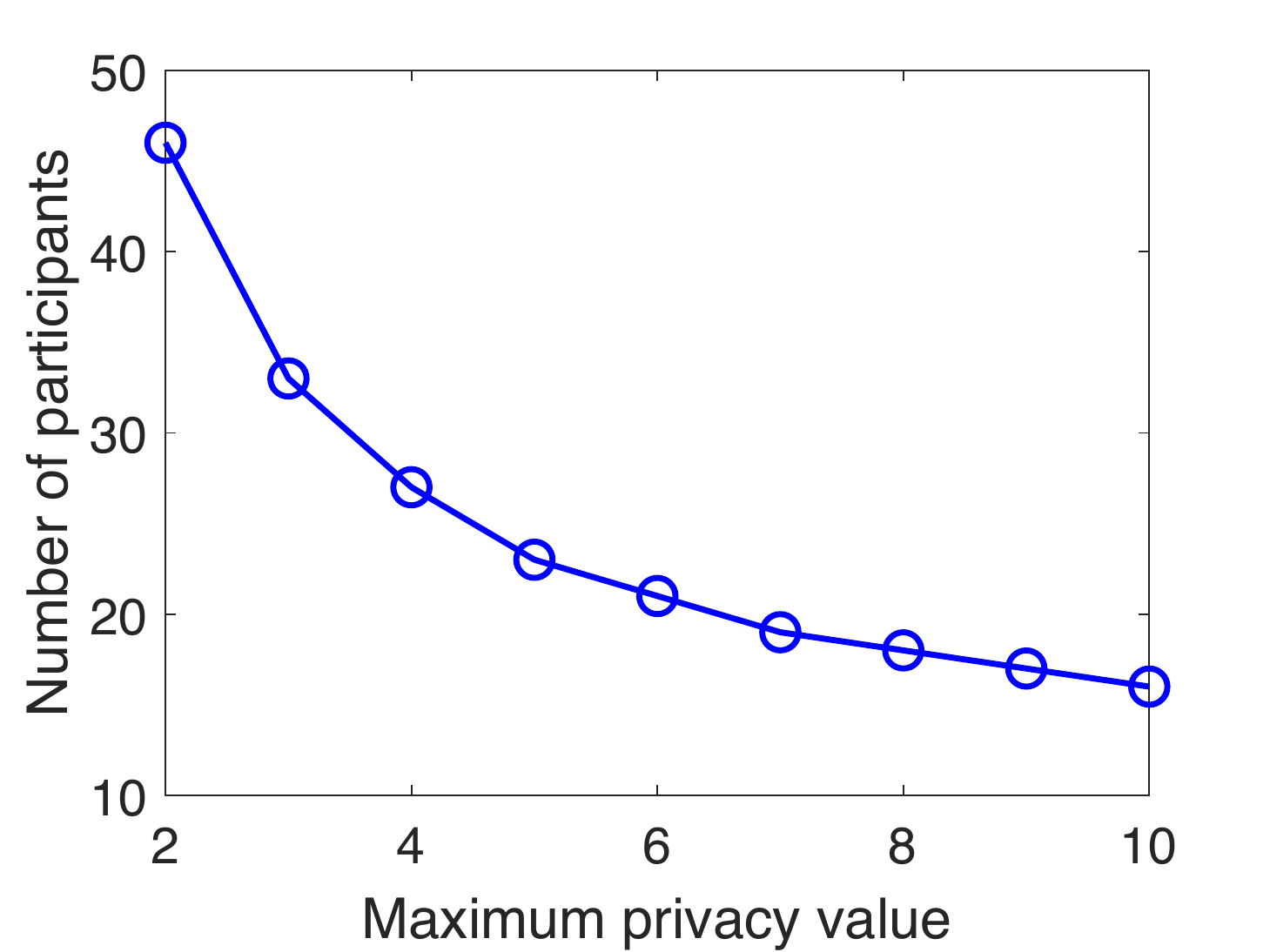}
  \caption{Impact of $\nu_{max}$.}
  \label{fig:v_max_num}
\end{subfigure}
\caption{The number of participating users under different settings of the number of users and the maximum privacy value.}
\label{fig:num}
\vspace*{-15pt}
\end{figure}

In our incentive mechanism, only users with positive privacy budgets will participate in the federated learning, i.e., if $i\notin\mathcal{S}$, user $i$ will not join the training. Therefore, we evaluate the size of $\mathcal{S}$ in our mechanism under different settings of $n$ and $\nu_{max}$. 
In Figure~\ref{fig:num}(\subref{fig:num_user_num}), we can see that as the number of users increases, the number of participants increases. This is reasonable because if there are more users interested in the federated learning task, more users will satisfy the condition of belonging to $\mathcal{S}$. In Figure~\ref{fig:num}(\subref{fig:v_max_num}), 
we can observe that as the privacy value of users becomes more and more diverse, the number of participants decreases since users with larger privacy value will not be chosen.

\subsection{Utility of Server}
\begin{figure}[ht]
\begin{subfigure}[t]{0.5\linewidth}
  \centering
  \includegraphics[width=\linewidth]{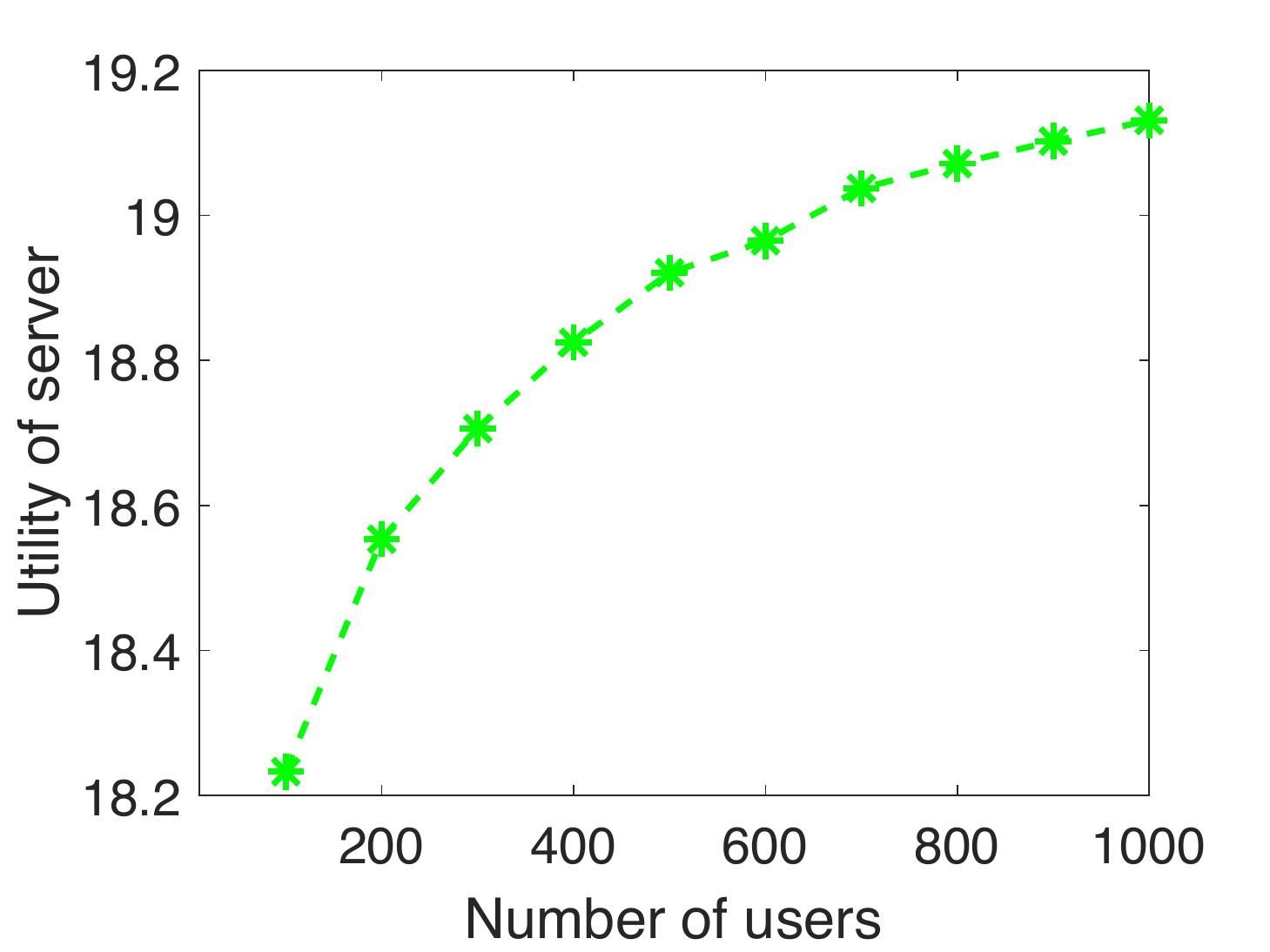}
  \caption{Impact of $n$.}
  \label{fig:num_user_u_s}
\end{subfigure}%
\begin{subfigure}[t]{0.5\linewidth}
  \centering
  \includegraphics[width=\linewidth]{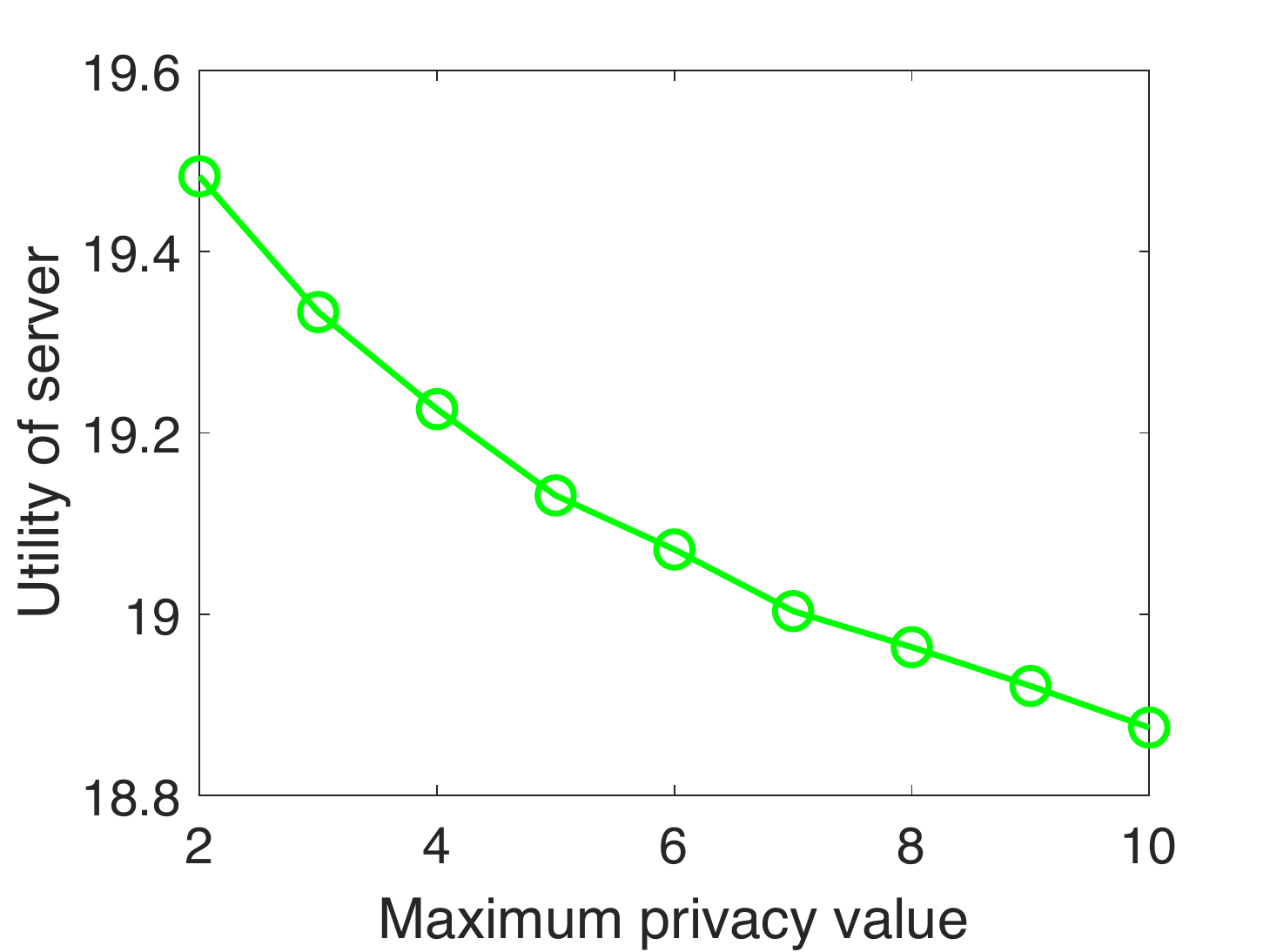}
  \caption{Impact of $\nu_{max}$.}
  \label{fig:v_max_u_s}
\end{subfigure}
\caption{Server's utility under different settings of the number of users and the maximum privacy value.}
\label{fig:u_s}
\vspace*{-12pt}
\end{figure}

We first evaluate the impact of the number of users on the server's utility. 
As shown in Figure~\ref{fig:u_s}(\subref{fig:num_user_u_s}), the utility of the server increases as the number of users increases. It is expected to see that the influence of the number of users on the server's utility is diminishing. 
With the results in Figure~\ref{fig:num}(\subref{fig:v_max_num}), it is reasonable that in Figure~\ref{fig:u_s}(\subref{fig:v_max_u_s}) the utility of the server decreases as the privacy value becomes more diverse.

\subsection{Utility of User}
\begin{figure}[ht]
\begin{subfigure}[t]{0.5\linewidth}
  \centering
  \includegraphics[width=\linewidth]{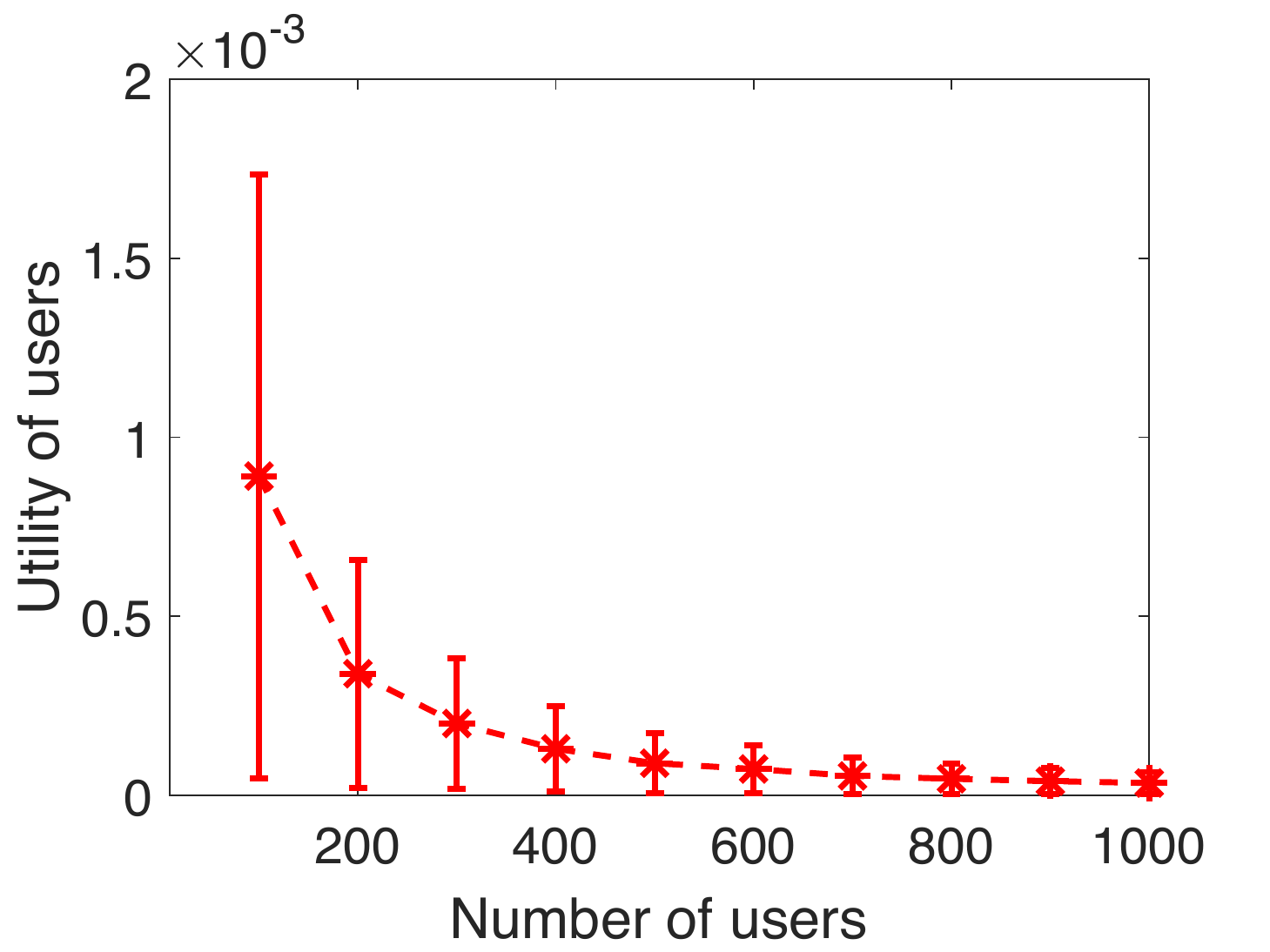}
  \caption{Impact of $n$.}
  \label{fig:num_user_u_i}
\end{subfigure}%
\begin{subfigure}[t]{0.5\linewidth}
  \centering
  \includegraphics[width=\linewidth]{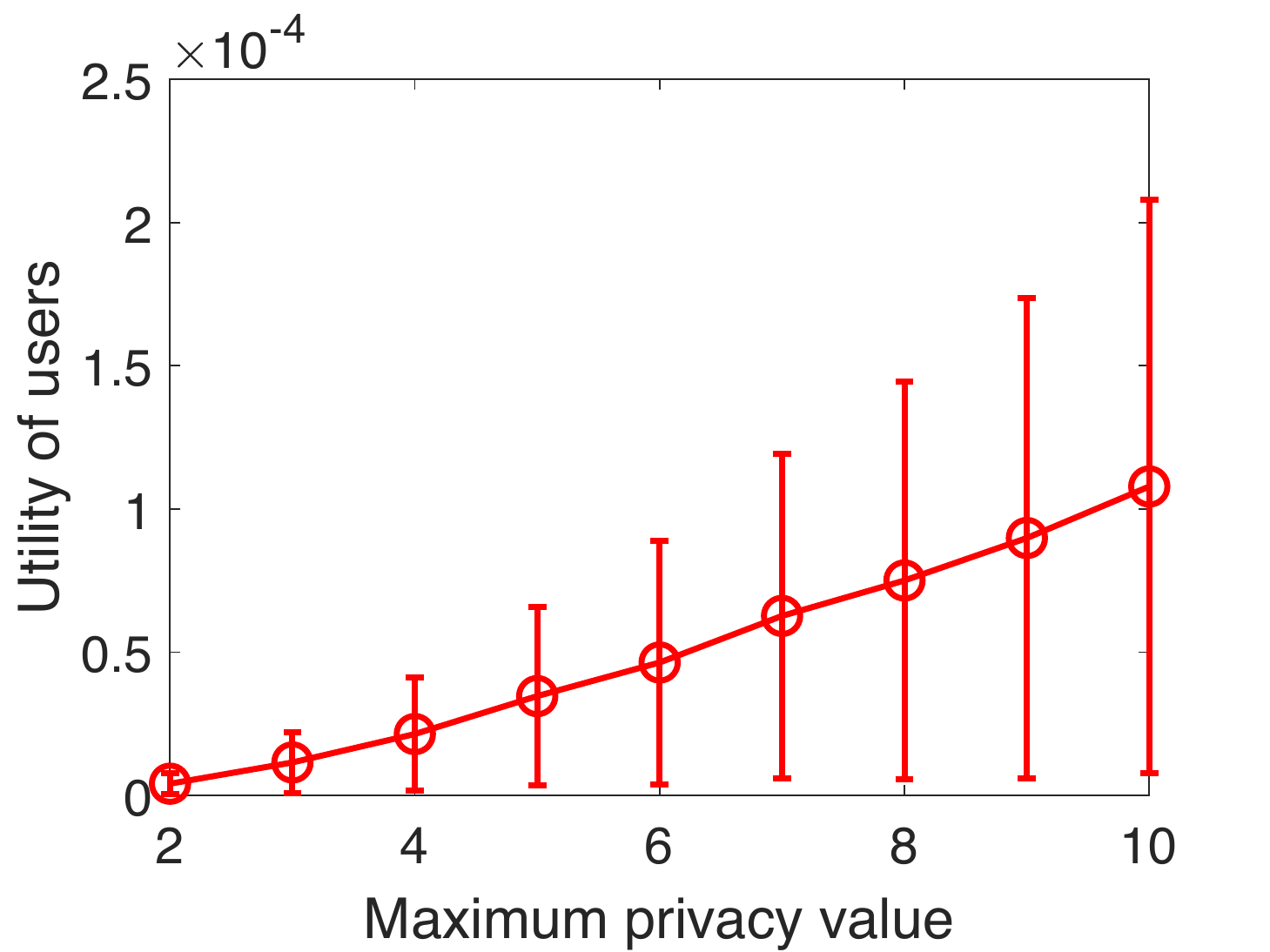}
  \caption{Impact of $\nu_{max}$.}
  \label{fig:v_max_u_i}
\end{subfigure}
\caption{The utility of users under different settings of the number of users and the maximum privacy value.}
\label{fig:u_i}
\vspace*{-12pt}
\end{figure}

We demonstrate users' utilities under different settings of $n$ and $\nu_{max}$ by showing the average utility of all users with the corresponding standard deviation. 
In Figure~\ref{fig:u_i}(\subref{fig:num_user_u_i}), the average and variance of users' utilities decrease with the number of users since more competitions are involved. In Figure~\ref{fig:u_i}(\subref{fig:v_max_u_i}), with the results in Figure~\ref{fig:num}(\subref{fig:v_max_num}), it is expected to see that users' utilities increases on average and becomes more diverse with the maximum privacy value.

\section{Conclusions}\label{sec:con}
In this paper, we have designed an incentive mechanism that can be used to motivate users with private data to participate in federated learning tasks. Our mechanism allows the cloud server to offer monetary rewards to compensate users for their privacy losses that occurs in the federated learning. We have adopted a two-stage Stackelberg game to model the utility maximization of the server and users. We have derived the best strategies for the server and users via solving the Stackelberg equilibrium. Through extensive numerical simulations, we have demonstrated the effectiveness of the mechanism. In future work, we plan to study the incentive design for federated learning tasks considering other costs of users, such as communication and computation cost.

\section*{Acknowledgment}
The work of R. Hu and Y. Gong was supported in part by the U.S. National Science Foundation under grants US CNS-2029685 and CNS-1850523. 

\ifCLASSOPTIONcaptionsoff
  \newpage
\fi
\bibliographystyle{IEEEtran}
\bibliography{hu}

\end{document}